%% file: iclr2027_conference.tex
\documentclass{article} % For LaTeX2e
\usepackage{iclr2027_conference,times}

\input{math_commands.tex}

\usepackage{hyperref}
\usepackage{url}
\usepackage{graphicx}
\usepackage{booktabs} 
\usepackage{multirow}
\usepackage{amssymb}
\usepackage{subcaption}
\usepackage{wrapfig}
\usepackage{adjustbox}

\makeatletter
\renewcommand\paragraph{\@startsection{paragraph}{4}{\z@}%
  {0.6ex plus 0.2ex minus 0.2ex}% 标题前的间距，改小这个值
  {-1em}%                          标题后到正文的水平间距
  {\normalsize\bfseries}}
\makeatother

\makeatletter
\g@addto@macro\normalsize{%
  \setlength\abovedisplayskip{4pt plus 2pt minus 2pt}%
  \setlength\belowdisplayskip{4pt plus 2pt minus 2pt}%
  \setlength\abovedisplayshortskip{0pt plus 2pt}%
  \setlength\belowdisplayshortskip{2pt plus 2pt}%
}
\makeatother
 
\title{CIRRA: Dual-Level Continual Instruction Reconciliation with Ongoing Execution for Embodied Robot Agents in Interactive Household Tasks}

\author{%
\makebox[\dimexpr\textwidth-2\tabcolsep\relax][c]{%
\begin{tabular}[t]{@{}c@{}}
{\normalfont\fontsize{11}{15}\selectfont\bfseries
Ci Zhang\textsuperscript{1}\quad
Enfu Nan\textsuperscript{2}\quad
Arman Akbari\textsuperscript{2}\quad
Lin Zhao\textsuperscript{2}\quad
Li Wang\textsuperscript{2}}\\[4pt]
{\normalfont\fontsize{11}{15}\selectfont\bfseries
Chen Wang\textsuperscript{3}\quad
Weiwei Chen\textsuperscript{3}\quad
Yanzhi Wang\textsuperscript{2}\quad
Geng Yuan\textsuperscript{1}}\\[10pt]
{\normalfont\small
\textsuperscript{1}\,University of Georgia}\\[2pt]
{\normalfont\small
\textsuperscript{2}\,Northeastern University}\\[2pt]
{\normalfont\small
\textsuperscript{3}\,EmbodyX Inc.}
\end{tabular}%
}%
}

\iclrfinalcopy
\begin{document}

\maketitle
\lhead{Preprint}

\begin{figure}[h]
    \centering
    \includegraphics[width=\textwidth]{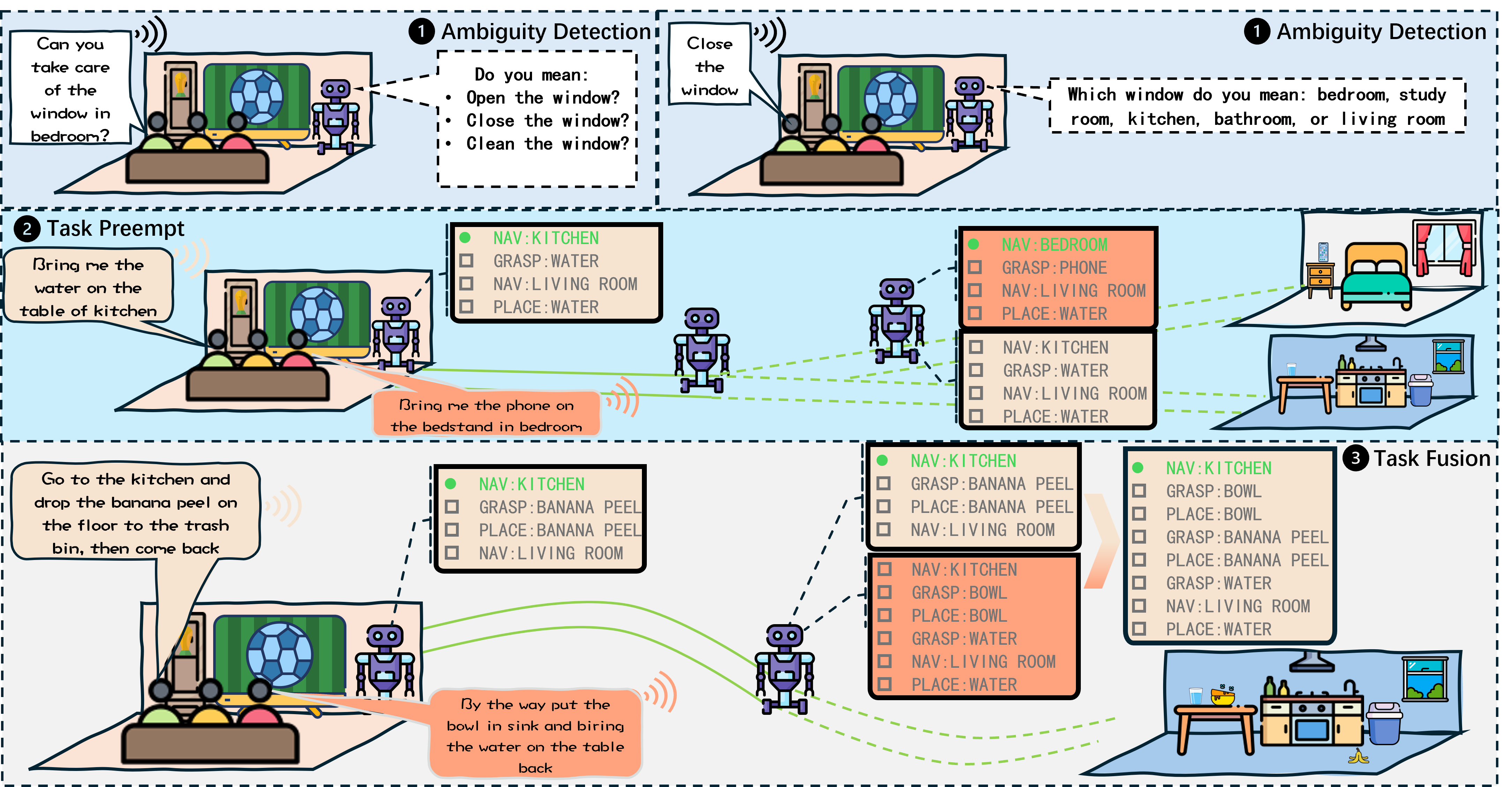}
    \caption{Demonstration of CIRRA's behaviors,          including (1) different types of instruction      ambiguity and their
        resolution, (2) task preemption, and (3) task fusion. The queue
        case, where an incoming instruction is directly scheduled for
        later execution, is omitted for brevity.}
    \label{fig:mainpic}
\end{figure}

\begin{abstract}
As the demand for assistance with household chores continues to grow, household robots have become a major focus of embodied intelligence. Real-world household deployment, however, inherently requires continuously replanning on human–robot interaction, as users may issue new instructions while previous requests are still under execution. Existing robot agents commonly accommodate such instructions by regenerating or extensively revising the remaining task sequence, which can introduce plan ambiguity, logical inconsistency, and redundant execution. To address these limitations, we formulate continual instruction reconciliation and propose CIRRA (Continual Instruction Reconciliation for Robot Agents), a dual-level framework combining LLM-based semantic reasoning with rule-constrained structural integration. CIRRA first uses semantic reasoning to determine whether an incoming instruction can be grounded to a unique executable skill and whether its required execution location is specified, resolving underspecified action and location information when necessary. It then preserves the ongoing subtask sequence as an execution backbone and generates fine-grained integration candidates by inserting the incoming subtasks into location-matched segments of the current task chain. Finally, the semantic reasoner evaluates only the modified segments to identify task dependencies and potential conflicts and select the most logically coherent local integration. By coupling structure-preserving local integration with semantic grounding and logical selection, CIRRA maintains alignment with the robot’s current execution, mitigates plan ambiguity and logical inconsistency, and reuses shared subtasks to reduce redundant execution. We further introduce CHIRP (Continual Household Instruction Reconciliation and Planning), a text-based benchmark of 120 episodes spanning eight household environments and six categories of everyday activities. On CHIRP, CIRRA reaches 74.2\% decision agreement, 30 points above the strongest replanning baseline, and every correct fusion decision it makes yields a correctly placed, conflict-free schedule. On a Unitree G1 humanoid, CIRRA interrupts ongoing skills at the correct moment in every trial and significantly outperforms all baselines on every metric.
\end{abstract}

\input{Tex/Introduction}
\input{Tex/related_work}
\input{Tex/Method}

\input{Tex/experiments}

\input{Tex/conclusion}

\section*{AI Use Statement}
Generative AI tools were used solely to improve the clarity and readability of the manuscript. All AI-assisted edits were reviewed and verified by the authors, who take full responsibility for the accuracy, originality, and integrity of the final manuscript.

\bibliography{iclr2027_conference}
\bibliographystyle{iclr2027_conference}

\appendix
\input{Tex/appendix}

\end{document}

%% file: math_commands.tex
\usepackage{amsmath,amsfonts,bm}

\def\eqref#1{equation~\ref{#1}}
\def\1{\bm{1}}

\DeclareMathAlphabet{\mathsfit}{\encodingdefault}{\sfdefault}{m}{sl}
\SetMathAlphabet{\mathsfit}{bold}{\encodingdefault}{\sfdefault}{bx}{n}

%% file: Tex/Introduction.tex
\section{Introduction}
Recently, household robots have emerged as a major focus in embodied intelligence, driven by the growing demand for everyday assistance amid the accelerating pace of modern life, population aging, and persistent labor shortages in care and domestic services. Recent advances in robotic foundation models, particularly vision-language-action models\cite{brohan2022rt,zitkovich2023rt,kim2024openvla,intelligence2025pi_,bjorck2025gr00t}, have made general-purpose household assistance increasingly feasible, enabling robots to understand natural-language instructions, perceive complex environments, and execute a diverse range of navigation and manipulation skills.

However, even capable action models alone remain insufficient for autonomous household operation \cite{shi2025hi,yang2025lohovla,koo2025hamlet}. To bridge this gap, embodied robot agents have extensively studied high-level task decomposition and long-horizon skill coordination \cite{dalal2024plan,yuan2025being}, task-state tracking and memory \cite{wang2025karma,koo2025hamlet}, and adaptive replanning in response to execution failures or environmental changes \cite{huang2022inner,skreta2024replan}. These advances enable an agent to translate a high-level user goal into a coherent sequence of executable skills, monitor task progress, and revise its plan when execution deviates from expectation.

Despite this progress, a critical challenge for household deployment remains comparatively underexplored: how to integrate a newly arriving task instruction with one that is already under execution. Although existing interactive agents can engage in dialogue and respond to user corrections, feedback, or task changes \cite{wei2022chain,kambhampati2024position,shinn2023reflexion,li2023interactive,joublin2024copal,borate2025llm,lu2025decompose}, a common strategy for accommodating such input is to re-invoke a high-level model with the updated instruction and current execution context to regenerate or revise the remaining task sequence. Without explicit constraints on preserving unaffected portions of the ongoing plan, this generative replanning can introduce three key issues. First, it may create \textbf{plan ambiguity}, as a minor instruction update can cause substantial plan churn, unnecessarily reorder pending subtasks, and obscure how the revised plan aligns with the current execution state. Second, it may produce \textbf{logical inconsistency} by violating task dependencies or introducing conflicting action orderings. Third, it may lead to \textbf{execution redundancy} by failing to identify and reuse subtasks shared between the ongoing and incoming instructions. These limitations highlight the need for fine-grained, execution-aligned instruction integration rather than unconstrained plan regeneration.

To address these limitations, we propose CIRRA, a dual-level framework that combines LLM-based semantic reasoning with rule-constrained structural integration. Upon receiving a new instruction, the semantic reasoning level first determines whether its action description can be grounded to a unique executable skill and whether the location required for execution is explicitly specified (shown in Ambiguity Detection in Fig.\ref{fig:mainpic}). It resolves underspecified action and location information to obtain a complete, location-aware representation of the incoming instruction. For a non-emergency instruction, the structural integration level then treats the robot’s ongoing subtask sequence as an immutable execution backbone(shown in Task fusion in Fig.\ref{fig:mainpic}): it retains every existing subtask and their relative ordering, and generates fine-grained integration candidates by inserting the incoming subtasks into segments of the sequence associated with the same execution location. The semantic reasoning level is subsequently invoked again, but only over the modified segments, to evaluate their task dependencies, identify potential conflicts, and select the most logically coherent local integration. In this way, CIRRA prevents the LLM from globally regenerating or reordering the ongoing plan, while still leveraging its semantic reasoning to accurately ground the incoming instruction and determine its logically appropriate placement. The two levels therefore provide complementary guarantees: rule-constrained integration preserves the robot’s current execution strategy, whereas LLM-based reasoning ensures instruction completeness, placement accuracy, and the logical coherence of each modified segment. Together, they enable fine-grained instruction reconciliation without disrupting ongoing execution or introducing redundant actions.

Existing household benchmarks primarily focus on instruction following, dialogue, and task execution under predefined user goals, but are not designed to evaluate replanning when new task requests arrive during ongoing execution \cite{shridhar2020alfred,padmakumar2022teach,gao2022dialfred}. Conversely, existing replanning benchmarks and evaluation settings largely study plan revision in response to execution failures or environmental changes, rather than continual user-driven task updates in interactive household settings. Consequently, there remains no dedicated benchmark at the intersection of household interaction and instruction-triggered replanning. To fill this gap, we introduce \textbf{CHIRP} (Continual Household Instruction Reconciliation and Planning), a text-based benchmark specifically designed to evaluate how embodied robot agents revise an ongoing household task plan in response to newly arriving user instructions. The benchmark spans eight representative household environments: bedroom, study room, kitchen, bathroom, living room, basement, garage, and yard. And covers six major categories of household activities: object fetching, cleaning, tidying, cooking, appliance operation, and waste disposal. Across all evaluated environments and task categories, CIRRA consistently outperforms existing replanning baselines, producing more accurate instruction integrations, more logically coherent task sequences, and less redundant execution.

%% file: Tex/related_work.tex
\section{Related Work}

\subsection{Robotic Action Foundation Models}

% With the emergence of a broad family of vision-language-action foundation models \cite{zitkovich2023rt,team2024octo,kim2024openvla,li2024cogact,intelligence2025pi_,qu2025spatialvla,bjorck2025gr00t,intelligence2025pi_,bu2025univla,shukor2025smolvla}, robot control has undergone a fundamental transition from separately trained, task-specific visuomotor policies toward reusable, language-conditioned action foundations capable of transferring knowledge across tasks, environments, and robot embodiments. By connecting the semantic knowledge of pretrained vision-language models with physical action generation, these models have substantially improved robots’ ability to understand natural-language instructions, recognize novel objects and task configurations, and adapt previously learned behaviors to new settings. Meanwhile, the evolution from autoregressive discrete-action prediction to specialized diffusion- and flow-based action modules has enabled more expressive continuous control and increasingly complex and dexterous manipulation. More recent developments have further incorporated explicit spatial representations, heterogeneous robot and human data, open-world co-training, humanoid embodiments, and efficient deployment. Together, these advances have transformed VLA models from isolated skill policies into increasingly general-purpose action interfaces that provide reusable execution capabilities for higher-level robotic systems.

Vision-language-action (VLA) foundation models~\cite{zitkovich2023rt,team2024octo,kim2024openvla,li2024cogact,intelligence2025pi_,qu2025spatialvla,bjorck2025gr00t,bu2025univla,shukor2025smolvla} have shifted robot control from separately trained, task-specific policies toward reusable, language-conditioned action models that transfer across tasks, environments, and embodiments. By grounding the semantic knowledge of pretrained vision-language models in continuous action generation, these models increasingly serve as general-purpose skill interfaces that higher-level robotic systems can invoke. As a result, the bottleneck of household autonomy moves to the higher level: deciding which skills to execute and in what order, especially when new instructions arrive during execution.

\subsection{Embodied Agents for Task Planning and Replanning}

At this higher level, LLM-based embodied agents translate user goals into executable robot behaviors. SayCan grounds LLM-proposed actions in the affordances of pretrained skills~\cite{ahn2022can}, ProgPrompt and Code as Policies generate program-like plans over available actions and control APIs~\cite{singh2022progprompt,liang2023code}, and Inner Monologue feeds success detection, scene descriptions, and human feedback back into planning~\cite{huang2022inner}.

Subsequent agents add hierarchical skill orchestration~\cite{dalal2024plan,yuan2025being}, persistent task memory~\cite{wang2025karma}, and online user feedback~\cite{shi2025hi}, and increasingly rely on replanning for robust long-horizon execution: CAPE corrects violated skill preconditions~\cite{raman2024cape}, AdaPlanner refines plans from environmental feedback~\cite{sun2023adaplanner}, RePLan revises behavior after plan failures or unforeseen obstacles~\cite{skreta2024replan}, and PRED pre-emptively revises upcoming actions when the environment changes~\cite{kim2024pre}. However, these methods predominantly treat replanning as recovery under a fixed user goal: the task itself stays unchanged, and the agent only adapts its behavior to execution-state mismatches.

This leaves an important case underexplored: in interactive households, users may issue additional task instructions while an earlier task chain is still being executed. Interactive agents incorporate corrections, feedback, or clarifications about the ongoing task~\cite{huang2022inner,shi2025hi}, but do not address how a new task should be reconciled with it. Simply prompting an LLM with the new instruction and the execution context gives no guarantee that unaffected subtasks and their order are preserved. A small update may thus cause \textbf{plan ambiguity}, where excessive plan churn obscures how the revised plan relates to the current execution state; \textbf{logical inconsistency}, where task dependencies are violated or actions are ordered in conflict; and \textbf{execution redundancy}, where navigation or manipulation subtasks shared by the ongoing and incoming instructions are not reused. To our knowledge, CIRRA is the first embodied-agent framework for continual, instruction-triggered replanning over an ongoing task chain. It combines execution-aligned structural integration with LLM-based semantic reasoning to preserve the unaffected part of the current plan while resolving all three issues.

%% file: Tex/Method.tex
\section{Methodology}
\label{sec:methodology}
\begin{figure*}[t]
    \centering
    \includegraphics[width=\textwidth]{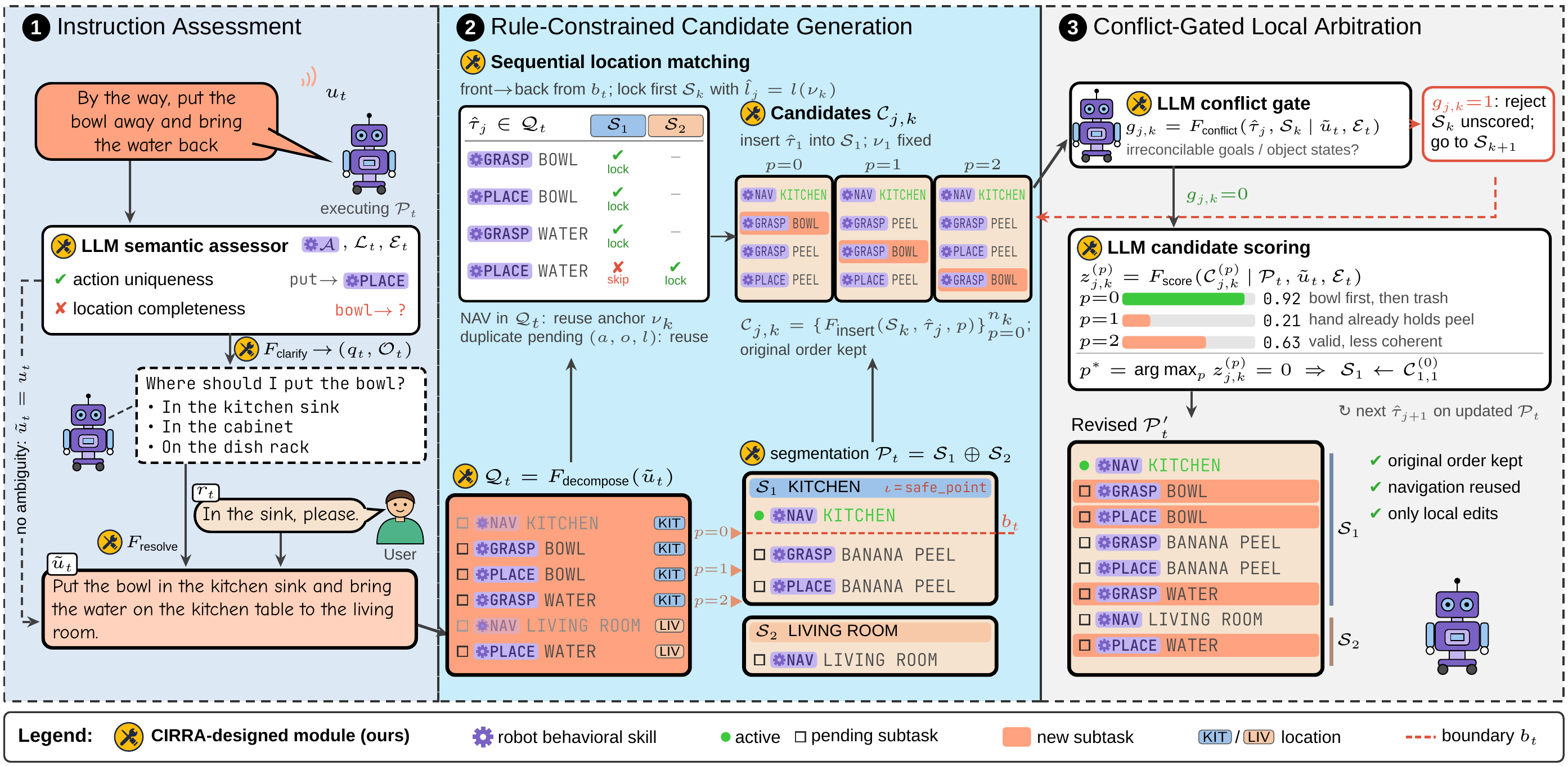}
    \caption{Overview of CIRRA: (1) instruction assessment and ambiguity resolution, (2) rule-constrained candidate generation within location-matched segments, and (3) conflict-gated LLM arbitration that yields the revised sequence $\mathcal{P}'_t$.}
\label{fig:method}
\end{figure*}

\subsection{Problem Formulation}

We consider an embodied robot agent that receives a new user
instruction while executing an existing household task. At time
step $t$, the agent maintains an ongoing task sequence
\begin{equation}
    \mathcal{P}_t
    =
    [\tau_1,\tau_2,\ldots,\tau_N],
    \label{eq:ongoing_sequence}
\end{equation}
where each subtask is represented using a unified tuple
\begin{equation}
    \tau_i
    =
    (a_i,o_i,l_i,s_i,\iota_i).
    \label{eq:subtask_representation}
\end{equation}
Here, $a_i$ denotes an executable robot skill. We divide the
robot skill set $\mathcal{A}$ into navigation skills
$\mathcal{A}_{\mathrm{nav}}$ and non-navigation behavioral
skills $\mathcal{A}_{\mathrm{beh}}$. The meanings of $o_i$ and
$l_i$ depend on the corresponding skill type. If
$a_i\in\mathcal{A}_{\mathrm{nav}}$, the subtask does not operate
on a target object, and we therefore set $o_i=\varnothing$.
In this case, $l_i$ denotes the destination of the navigation
subtask. If $a_i\in\mathcal{A}_{\mathrm{beh}}$, $o_i$ denotes
the object on which the behavior is performed, while $l_i$
denotes the location at which the behavior should be executed. Thus,
\begin{equation}
    o_i=\varnothing
    \quad\text{if}\quad
    a_i\in\mathcal{A}_{\mathrm{nav}}.
    \label{eq:navigation_object}
\end{equation}
The variable $s_i$ denotes the execution status of the subtask:
\begin{equation}
    s_i
    \in
    \{
    \texttt{pending},
    \texttt{active},
    \texttt{done}
    \}.
    \label{eq:execution_status}
\end{equation}
A $\texttt{pending}$ subtask has not yet started,
$\texttt{active}$ denotes the subtask currently being executed,
and $\texttt{done}$ indicates that the subtask has already been
completed.

The variable $\iota_i$ denotes the interruptibility mode of the
subtask:
\begin{equation}
    \iota_i
    \in
    \{
    \texttt{immediate},
    \texttt{safe\_point},
    \texttt{uninterruptible}
    \}.
    \label{eq:interruptibility_mode}
\end{equation}
Specifically, $\texttt{immediate}$ permits the subtask to be
interrupted at any time. The $\texttt{safe\_point}$ mode permits
interruption only after the subtask reaches its next predefined
safe execution boundary. The $\texttt{uninterruptible}$ mode
requires the subtask to finish before an incoming instruction
can affect ongoing execution. These modes apply to both
navigation and behavioral subtasks and prevent plan integration
from interrupting a robot skill at an invalid or unsafe state.

While executing $\mathcal{P}_t$, the robot receives an
additional natural-language instruction $u_t$. Given the ongoing
task sequence $\mathcal{P}_t$ and the incoming instruction
$u_t$, our objective is to construct a revised task sequence
$\mathcal{P}'_t$ that accommodates the new request while
preserving the unaffected subtasks and execution structure of
$\mathcal{P}_t$.

\subsection{Instruction Assessment and Ambiguity Resolution}

Before integrating an incoming instruction with the ongoing
task sequence, CIRRA first assesses whether the instruction can
be unambiguously grounded into executable robot subtasks (Figure~\ref{fig:method},1). Given
the incoming instruction $u_t$, the LLM-based semantic assessor
is provided with the robot's executable skill set $\mathcal{A}$,
the available household locations $\mathcal{L}_t$, and the
current environment context $\mathcal{E}_t$.

The assessor examines every action expression in $u_t$ along two
dimensions: action uniqueness and location completeness. An
action expression is considered ambiguous if it cannot be
uniquely mapped to one skill in $\mathcal{A}$. Its location
information is considered incomplete if the location required
for executing the corresponding action is omitted or cannot be
uniquely determined from $\mathcal{L}_t$ and
$\mathcal{E}_t$.
\begin{equation}
    (q_t,\mathcal{O}_t)
    =
    F_{\mathrm{clarify}}
    (u_t,\mathcal{A},\mathcal{L}_t,\mathcal{E}_t),
    \label{eq:clarification_generation}
\end{equation}
When an action or execution location is underspecified,
CIRRA does not autonomously infer the missing information.
Instead, $F_{\mathrm{clarify}}$ generates a targeted
clarification question $q_t$ together with a compact set of
contextually relevant candidate interpretations
$\mathcal{O}_t$. These candidates are constrained by the
robot's executable skill set $\mathcal{A}$, the available
household locations $\mathcal{L}_t$, and the current
environment context $\mathcal{E}_t$. The robot then presents
$q_t$ and $\mathcal{O}_t$ to the user and obtains a response $r_t$. 
\begin{equation}
    \tilde{u}_t
    =
    F_{\mathrm{resolve}}
    (u_t,r_t,\mathcal{A},\mathcal{L}_t,\mathcal{E}_t),
    \label{eq:ambiguity_resolution}
\end{equation}
Based on this response $r_t$, $F_{\mathrm{resolve}}$ produces
a normalized instruction $\tilde{u}_t$ in which every action is
associated with a unique executable skill and an explicit
execution location.

If no action or location ambiguity is detected, the
clarification interaction is skipped and
$\tilde{u}_t=u_t$. The resulting normalized instruction
$\tilde{u}_t$ is then passed to the task-decomposition and
plan-integration stage described next.

\subsection{Rule-Constrained Execution-Aligned Candidate Generation}

After obtaining the normalized instruction $\tilde{u}_t$, CIRRA
decomposes it into an ordered sequence of executable incoming subtasks (Figure~\ref{fig:method},2):
\begin{equation}
    \mathcal{Q}_t
    =
    F_{\mathrm{decompose}}(\tilde{u}_t)
    =
    [\hat{\tau}_1,\hat{\tau}_2,\ldots,\hat{\tau}_M].
    \label{eq:incoming_subtasks}
\end{equation}
Each incoming subtask is represented as
$\hat{\tau}_j=(\hat{a}_j,\hat{o}_j,\hat{l}_j,
\hat{s}_j,\hat{\iota}_j)$, where
$\hat{s}_j=\texttt{pending}$. The sequence $\mathcal{Q}_t$
contains only the subtasks derived from the incoming instruction
and has not yet been integrated with $\mathcal{P}_t$.

CIRRA first determines the earliest point at which the incoming
instruction may affect ongoing execution. Let $\tau_{k_t}$ be
the currently active subtask. If
$\iota_{k_t}=\texttt{immediate}$, it may be paused immediately
and resumed after the inserted subtasks are completed. If
$\iota_{k_t}=\texttt{safe\_point}$, integration is delayed
until the next safe execution boundary. If
$\iota_{k_t}=\texttt{uninterruptible}$, the active subtask must
finish before the revised sequence can take effect. We denote
the resulting earliest admissible integration boundary by
$b_t$; execution before $b_t$ remains fixed.

\paragraph{Navigation-anchored segmentation.}
The ongoing sequence $\mathcal{P}_t$ is organized into
navigation-anchored segments. Every navigation subtask begins a
new segment, and all subsequent non-navigation subtasks are
assigned to that segment until the next navigation subtask:
\begin{equation}
    \mathcal{P}_t
    =
    \mathcal{S}_1\oplus\cdots\oplus\mathcal{S}_K,
    \qquad
    \mathcal{S}_k
    =
    [\nu_k,\tau_{k,1},\ldots,\tau_{k,n_k}].
    \label{eq:navigation_segments}
\end{equation}
Here, $\nu_k$ is the leading navigation subtask, and
$l(\mathcal{S}_k)=l(\nu_k)$ defines the execution location of
the segment. If the robot visits the same location multiple
times, each navigation subtask starts a distinct segment,
preserving the temporal structure of the original plan.

\paragraph{Sequential matching and candidate enumeration.}
CIRRA processes every non-navigation subtask
$\hat{\tau}_j\in\mathcal{Q}_t$ individually. Beginning from the
earliest segment permitted by $b_t$, it traverses
$\mathcal{P}_t$ from front to back and compares
$\hat{l}_j$ with the destination $l(\nu_k)$ of each segment.
A segment is skipped when their locations differ. Once
$\hat{l}_j=l(\nu_k)$, $\mathcal{S}_k$ is temporarily locked as
the current insertion segment.

CIRRA keeps $\nu_k$ fixed and enumerates every position at which
$\hat{\tau}_j$ can be inserted among the non-navigation
subtasks:
\begin{equation}
    \mathcal{C}_{j,k}
    =
    \left\{
    F_{\mathrm{insert}}
    (\mathcal{S}_k,\hat{\tau}_j,p)
    \mid p=0,\ldots,n_k
    \right\}.
    \label{eq:local_candidates}
\end{equation}
The case $p=0$ places $\hat{\tau}_j$ immediately after
$\nu_k$, while $p=n_k$ places it after all existing behavioral
subtasks. Every candidate preserves the leading navigation
subtask and the relative ordering of all original subtasks.
Thus, integrating a behavior into a location-matched segment
reuses the segment's navigation instead of introducing another
navigation action.

Before enumeration, CIRRA also checks whether the matched
segment already contains an active or pending subtask with the
same action, object, and location. If so, the existing subtask
is reused and no duplicate is inserted. Otherwise,
$\mathcal{C}_{j,k}$ is passed to the LLM judge. Consequently,
all modifications remain local, while the original subtasks and
their relative ordering are strictly preserved.

\subsection{LLM-Based Conflict-Gated Local Arbitration}

For each location-matched segment, CIRRA applies a
conflict-first arbitration procedure (Figure~\ref{fig:method},3). The LLM first determines
whether the incoming action has an irreconcilable conflict with
any behavioral action in $\mathcal{S}_k$:
\begin{equation}
    g_{j,k}
    =
    F_{\mathrm{conflict}}
    (\hat{\tau}_j,\mathcal{S}_k
    \mid\tilde{u}_t,\mathcal{E}_t),
    \qquad
    g_{j,k}\in\{0,1\}.
    \label{eq:conflict_gate}
\end{equation}
Here, $g_{j,k}=1$ indicates a conflict, while $g_{j,k}=0$
indicates compatibility. The conflict gate focuses on
incompatibilities that cannot be resolved by changing the
insertion position, such as mutually exclusive goals or
contradictory object-state requirements. Ordering dependencies
that can be resolved through an appropriate placement are
handled during candidate scoring.

If $g_{j,k}=1$, CIRRA rejects the entire candidate segment
without scoring its insertion positions. The traversal then
continues from $\mathcal{S}_{k+1}$. This avoids evaluating
candidate orders that cannot produce a valid integration
regardless of where the incoming action is inserted.

If $g_{j,k}=0$, the LLM scores every candidate in
$\mathcal{C}_{j,k}$:
\begin{equation}
    z^{(p)}_{j,k}
    =
    F_{\mathrm{score}}
    (\mathcal{C}^{(p)}_{j,k}
    \mid\mathcal{P}_t,\tilde{u}_t,\mathcal{E}_t),
    \qquad
    p=0,\ldots,n_k.
    \label{eq:candidate_score}
\end{equation}
The score measures whether the insertion preserves causal
dependencies, valid object-state transitions, and a coherent
local action ordering. CIRRA then selects the highest-scoring
position:
\begin{equation}
    p^{*}
    =
    \underset{p\in\{0,\ldots,n_k\}}{\arg\max}
    \;z^{(p)}_{j,k},
    \qquad
    \mathcal{S}_k
    \leftarrow
    \mathcal{C}^{(p^{*})}_{j,k}.
    \label{eq:best_local_insertion}
\end{equation}
Once an insertion is accepted, CIRRA terminates the traversal
for $\hat{\tau}_j$ and does not examine later segments, even if
they share the same location. If the current segment is rejected
by the conflict gate, traversal continues until another
location-matched segment is found. The next incoming behavioral
subtask is processed using the updated task sequence.

In this way, deterministic rules control the admissible segment
and insertion positions, while the LLM only rejects conflicts
and selects among local candidates. CIRRA therefore preserves
the unaffected execution structure, reuses existing navigation,
and prevents the LLM from globally regenerating or reordering
the ongoing task sequence.

%% file: Tex/experiments.tex
% ============================================================================
% EXPERIMENTS (full draft, 2026-07-29). All numbers match the frozen tables
% (Tables 1-4 generators under benchmark/). No em-dashes. Plain language.
% Citation keys assume the same bib entries as the tables.
% ============================================================================
\section{Experiments}

\subsection{Task Representation and Benchmark}

\paragraph{Benchmark.}
CHIRP contains 120 episodes set in a simulated home with eight rooms, covering six action categories of 20 episodes each (appliance control, cleaning, tidying, cooking, fetch and delivery, and waste handling). Each episode freezes the moment a new utterance arrives during an ongoing task and asks for the agent's reaction. Episodes are built as minimal pairs with nearly identical wording but different execution states, so a method that ignores execution state must miss one episode of each pair. The gold decisions comprise 40 fuse, 47 queue, 21 clarify and 12 preempt episodes, and each utterance comes in three surface forms. Every episode also provides a textual world snapshot (positions, object placements and held items), giving each baseline the environmental context it expects. Gold labels follow the decision doctrine of Section~\ref{sec:methodology}; the full schema, templates and annotation records are given in Appendix \ref{app:benchmark}.

\paragraph{Evaluation Metrics}
\label{sec:metrics}

We report three metrics.
\textbf{DA (decision agreement)} is the fraction of episodes whose response falls into the gold decision class (fuse, preempt, queue, or clarify).
\textbf{TA (target agreement)} is the fraction of fuse-gold episodes in which a method fuses and places the new subtasks at the gold position without moving any old subtask.
\textbf{EA (execution agreement)} is the strictest metric: the fraction of fuse-gold episodes in which, in addition, the whole schedule replays without conflict in the symbolic simulator and the new subtasks follow the gold order within their segment.

\input{Tables/main_results_table}

\paragraph{Evaluation Protocol}
We evaluate two conditions. In the primary \textbf{end-to-end}
condition, methods receive the raw utterance and episode state,
with all plan-consuming methods sharing a single cached Qwen3-8B
compilation. The \textbf{oracle spec} condition replaces this
compilation with the ground-truth specification to isolate
scheduling from language understanding. All LLM calls use identical
decoding settings and schema-constrained JSON output.
Evaluation is deterministic and uses no LLM judges.

\subsection{Baselines}

% Every compared system is coupled to its own execution stack, so none can
% run our episodes directly. ITP, CoPAL, CoMuRoS, DPPM, and SayCan are
% therefore policy-level reproductions: we reimplement each system's
% published reaction policy inside our harness, keeping the original prompt
% where one is available (CoPAL from the template files in the authors'
% repository, DPPM from the merge template quoted in its paper, the rest
% from the papers' descriptions). CoT, Reflexion, and LLM-Modulo are
% strategy families implemented from their publications, with LLM-Modulo
% using our symbolic simulator as its verifier. SayCan and the memory agent
% also quantify the sequential one-request contract of household service
% agents. LLM-direct reads a text rendering of the state plus the new utterance
% and outputs one of the four decisions directly, with no schedule.
% Conjunctive replan hands the state, the old plan, and the new request to
% the LLM and regenerates the whole schedule in one shot. Ours w/o judge
% is our full system with the LLM judge disabled: nothing else changes,
% but no candidate is ever vetoed by a compatibility question. Per-baseline fidelity and every adaptation are itemized in
% Appendix~B.

We implement policy-level reproductions of ITP, CoPAL, CoMuRoS,
DPPM, and SayCan in a shared harness, retaining original prompts
where available. CoT, Reflexion, and LLM-Modulo follow their
published strategies; LLM-Modulo uses our symbolic simulator as
its verifier. SayCan and the memory agent serve as
sequential-request baselines. LLM-direct predicts one of the four
decisions from the state and utterance without generating a
schedule. Conjunctive replan regenerates the full schedule from
the state, existing plan, and new request. Ours w/o judge disables
the LLM compatibility judge. Prompt sources and per-baseline
adaptations are detailed in Appendix \ref{app:baselines}.

\subsection{Results and Analysis}

% \paragraph{GPT Output}
% Our full system achieves 74.2\% DA
% (Table~\ref{tab:main-results}), compared with 59.2\% for
% LLM-direct and 44.2\% for Reflexion, the strongest
% schedule-generating baseline. DA rises to 93.0\% on the
% 43 compile-correct episodes. Of 31 end-to-end errors,
% 28 stem from compilation and three from false carry vetoes
% by the judge; the scheduling mechanism achieves 100\% DA
% under oracle specifications. Despite imperfect compilations,
% 49/77 episodes (64\%) still yield the correct decision.
% Regeneration baselines violate prior commitments in up to
% 84/120 episodes, whereas our method preserves them by
% construction.

% \paragraph{Claude Output}
Table~\ref{tab:main-results} summarizes the results. CIRRA reaches 74.2\% DA, 30 points above the strongest replanning baseline (Reflexion, 44.2\%) and 15 points above LLM-direct (59.2\%). Its remaining errors stem almost entirely from compilation: the shared Qwen3-8B compiles only 35.8\% of utterances exactly, and on these 43 episodes CIRRA reaches 93.0\%. Of its 31 errors, 28 originate in compilation and 3 are incorrect conflict rejections by the LLM judge; under oracle specifications CIRRA scores 100\%. The rule-constrained stages also absorb compilation errors: 49 of the 77 imperfectly compiled episodes (64\%) still yield the gold decision, and the majority of the 31 errors end in a clarification question rather than a wrong action. DA also understates the gap, because regeneration baselines break prior commitments, dropping or reordering pending steps in up to 84 of 120 episodes (CoMuRoS; CoPAL 51, DPPM 39; Appendix \ref{app:baselines}), whereas CIRRA only inserts into the existing sequence and cannot violate a commitment by design.

\input{Tables/metrics_table}
\paragraph{Comparison on fusion episodes.}
Table~\ref{tab:metric-suite} evaluates fusion, the core capability of CIRRA, on the 40 fuse-gold episodes. Five baselines never fuse. This includes LLM-Modulo: its verifier removes all commitment violations, yet the method never chooses to fuse, since a verifier can ensure that a plan is valid but not that it is the right reaction. The strongest baselines, ITP and DPPM, choose to fuse in 50.0\% and 45.0\% of the episodes, but only 15.0\% and 12.5\% of the episodes end in a fully correct schedule (EA). CIRRA fuses in 75.0\% of the episodes and every schedule it fuses is correct, reaching 75.0\% EA; without the judge, EA drops to 47.5\%, showing that the judge removes conflicting or misordered insertions. With oracle specifications, CIRRA reaches 100\% EA, whereas the best replanning baseline reaches only 40.0\%, showing that the gap comes from how methods react rather than from language understanding.

\paragraph{Effect of the judge.}
Comparing the two Ours rows isolates the LLM judge. Beyond the fusion gain above, its effect varies across categories (Table~\ref{tab:main-results}): it improves waste handling by 40 points by rejecting forbidden joint transports and falling back to queueing, but lowers fetch and delivery by 30 points because the 8B judge also rejects valid multi-item deliveries. Since all other components are rule-based, these errors reflect the capability of the judge model, which the scale ablation examines (Table~\ref{tab:scale-ablation}).

\input{Tables/scale_ablation_table}
\paragraph{Scale moves compilation, not the mechanism.}
Table~\ref{tab:scale-ablation} swaps the compiler and judge together from
Qwen3-4B to 8B then to 32B under the identical contract. Compile fidelity rises from
31.7\% to 35.8\% and then to 68.3\% and end-to-end DA tracks it from 65.8\% to 74.2\% and then to 91.6\%, while
the oracle condition holds at 100\% at every scale: the bottleneck is
language, the mechanism is scale-invariant, and gains from larger models
accrue to our system directly.

% =====================================================================
%  Real-World Deployment -- MAIN TEXT (Tex/real_world.tex)
%  \input{Tex/real_world} at the end of experiments.tex.
%  Details live in Tex/real_world_appendix.tex (\input after \appendix).
%  Placeholders are \ph{...}; search for "\ph{" before submission.
% =====================================================================
\providecommand{\ph}[1]{\textbf{[#1]}}
\providecommand{\sk}[1]{\texttt{#1}}

\subsection{Real-World Deployment}
\label{sec:real_world}

\begin{figure*}[t]
  \centering
  \begin{minipage}[b]{0.346\textwidth}
    \centering
    \includegraphics[width=\linewidth]{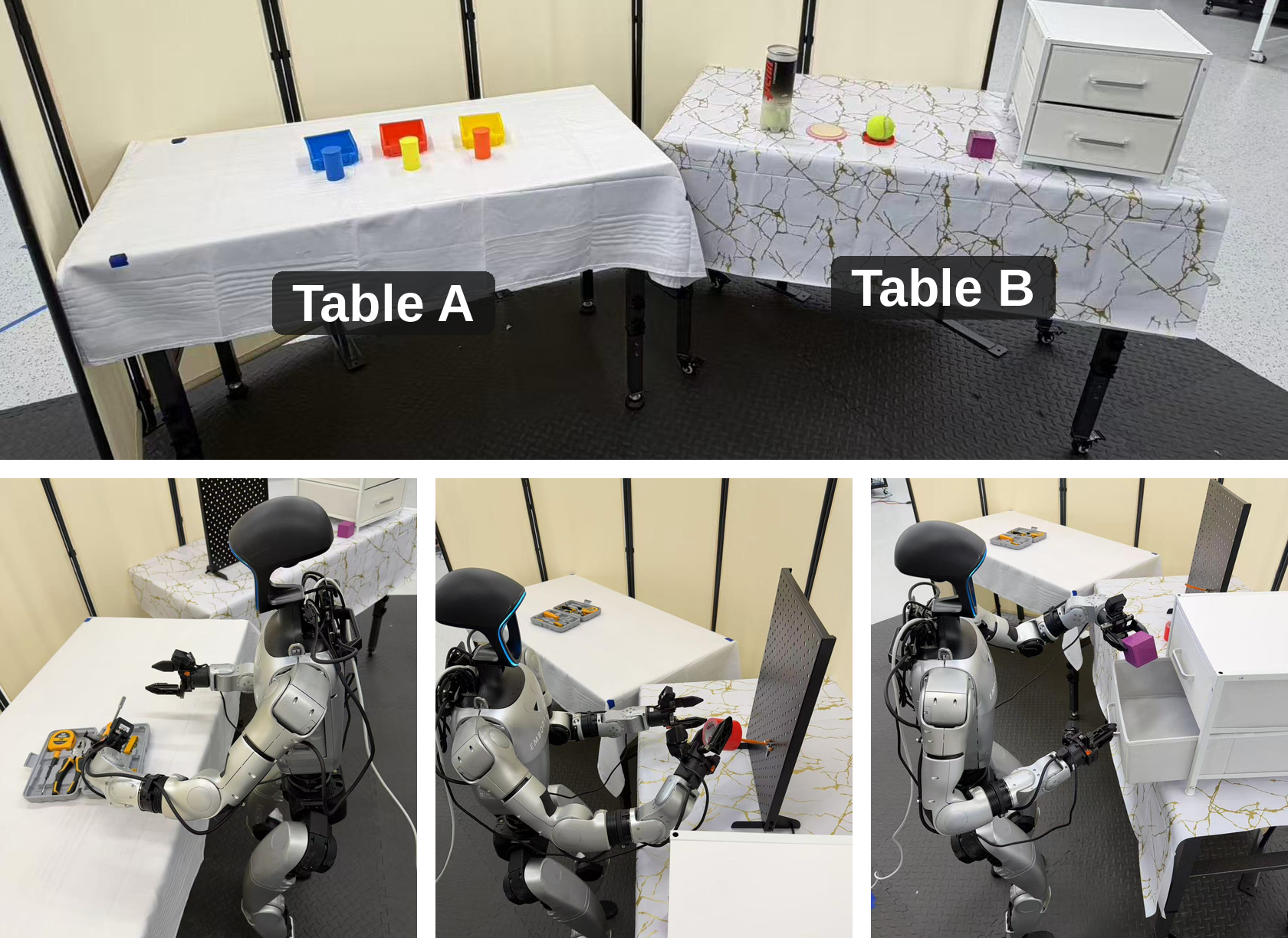}\\[2pt]
    {\small (a) Deployment environment}
  \end{minipage}\hfill
  \begin{minipage}[b]{0.641\textwidth}
    \centering
    \includegraphics[width=\linewidth]{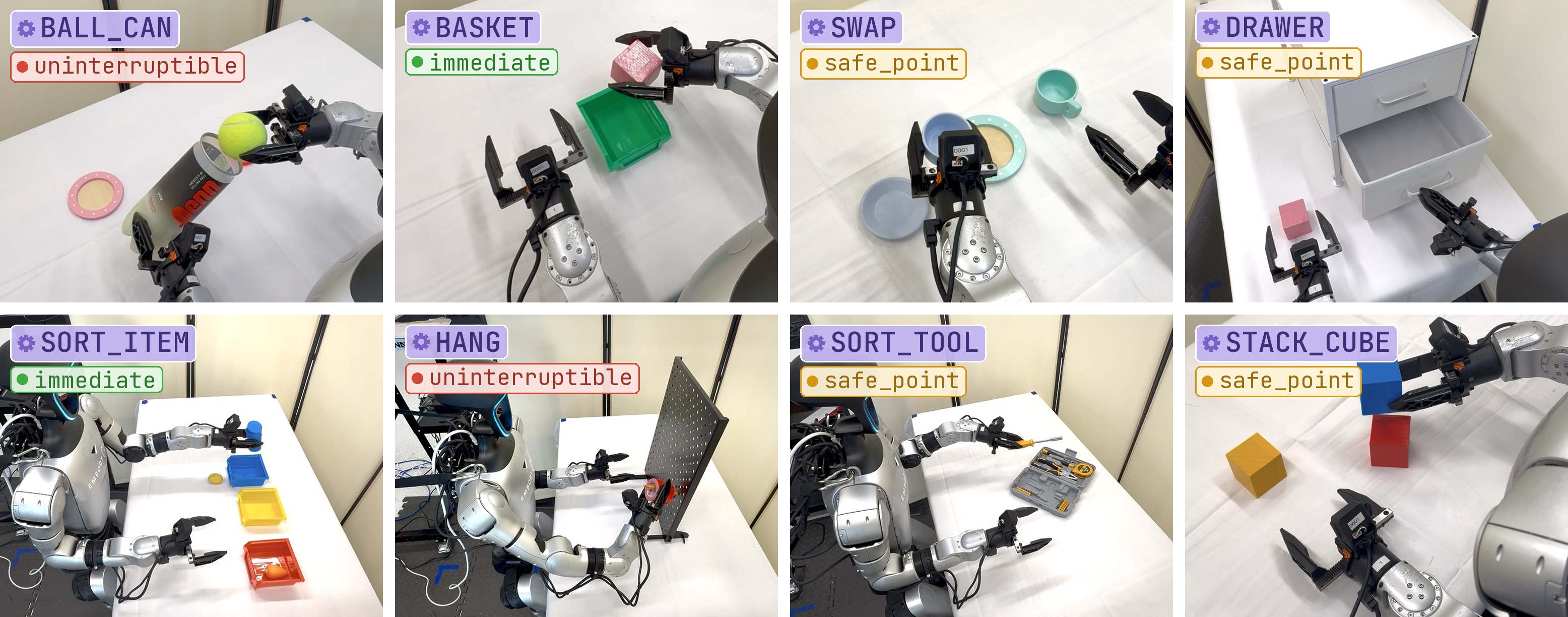}\\[2pt]
    {\small (b) Eight manipulation skills labeled with its interruptibility mode.}
  \end{minipage}
  \caption{Real-world setup. (a) A Unitree G1 humanoid operates at table $A$ (left) and table $B$ (right);(b) The eight tasks forming the robot's behavioral skills.}
  \label{fig:real-robot}
\end{figure*}

\begin{figure*}[t]
    \centering
    \includegraphics[width=\textwidth]{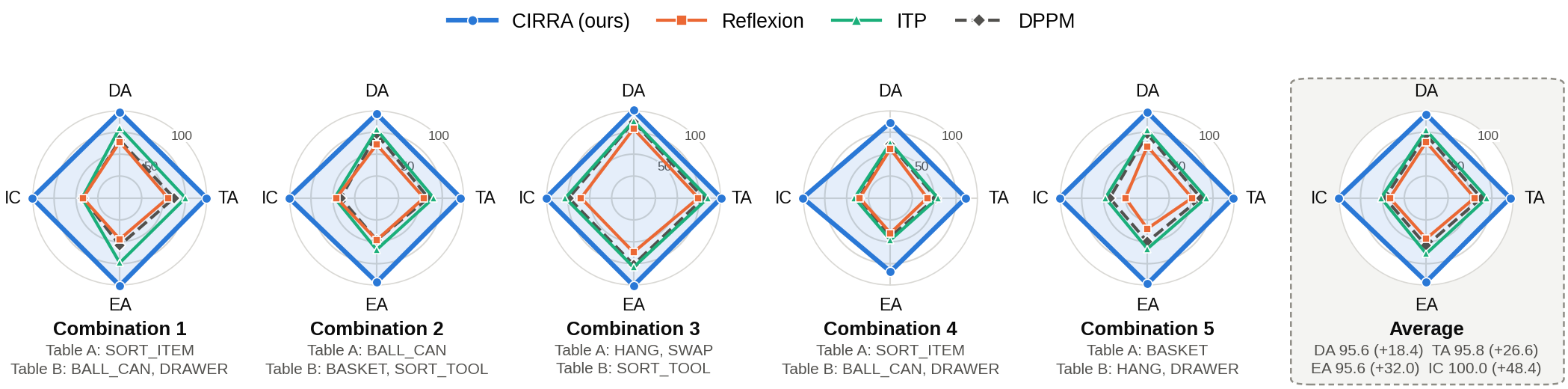}
    \caption{Real-world results (\%). The first five radars show the five task combinations; the rightmost averages them and lists CIRRA's mean on each metric, with its gain over the best baseline in parentheses.}
    \label{fig:real_radar}
\end{figure*}

% IC measures whether the agent respects the interruptibility of the active subtask

On a physical robot, a wrong reconciliation decision has physical
consequences: interrupting a skill at an invalid state can drop a held
object, and appending a request that could have been fused costs an
extra trip between locations. We therefore evaluate CIRRA on a real
robot.

\paragraph{Setup.}
A Unitree G1 humanoid operates at two side-by-side tables, $A$ and $B$ (shown in (a) of Fig \ref{fig:real-robot}) , walking between them with its built-in locomotion
controller (\sk{NAV}). Its behavioral skills are eight 
policies, each fine-tuned from ImageWAM~\cite{zhang2026imagewam} on 50
demonstrations of one task.
% TODO: replace \ph{citation} with \citep{...}
Each skill is assigned one of the three interruptibility modes of Section~\ref{sec:methodology} (\texttt{immediate}, \texttt{safe\_point}, or \texttt{uninterruptible}); in our setup, \sk{NAV} is always \texttt{immediate}, and the mode of each manipulation skill is listed in Appendix \ref{app:real}. Images from the onboard RGB-D camera
are passed directly to the Qwen3-VL-8B as $\mathcal{E}_t$.

% \paragraph{Task combinations.}
% Each combination places three tasks on the two tables
% (Figure~\ref{fig:real_radar}) and targets one behavior: preemption while
% an \texttt{immediate}~(1) or an \texttt{uninterruptible}~(2) skill is
% active, queueing~(3), fusion into a pending segment beyond a navigation
% step~(4), and fusion whose order within the segment must be inferred
% from the perceived layout~(5). The new instruction $u_t$ is issued at
% different phases of execution, e.g., during navigation, at the onset of
% a grasp or mid-manipulation, so the gold integration boundary $b_t$
% varies across trials. Each combination is run 25 times (125 episodes);
% instructions and gold sequences are listed in
% Appendix~\ref{app:real_scenarios}.
% % TODO: with 25 trials per combination every per-combination percentage
% % should be a multiple of 4; state the actual denominator if it differs.

\paragraph{Task combinations.}
Each combination consists of three tasks placed on the two tables (listed below each radar in Figure~\ref{fig:real_radar}) and targets one of three decisions: preemption (combinations 1 and 2), queueing (3), or fusion (4 and 5). Each combination is tested 25 times (125 trials in total), excluding trials in which a manipulation policy fails to grasp. In each trial, the new instruction arrives at a different point of the ongoing task sequence, so it may arrive while an \texttt{immediate}, \texttt{safe\_point}, or \texttt{uninterruptible} subtask is active. Instructions and gold sequences are at Appendix \ref{app:real}.

\paragraph{Baselines and metrics.}
We compare with the replanning baselines Reflexion, ITP and DPPM, none of which accounts for the interruptibility of the active subtask.. All methods are queried on identical execution-state snapshots and executed with the same skills. Besides DA, TA and EA, we report \emph{interruption compliance} \textbf{(IC)}, the fraction of episodes in which the new instruction interrupts the active subtask at the correct moment given its interruptibility mode: immediately for an \texttt{immediate} subtask, at its next safe point for a \texttt{safe\_point} subtask, and only after it finishes for an \texttt{uninterruptible} subtask.

\paragraph{Results.}
Averaged over the five combinations, CIRRA reaches 95.6\% DA, 95.8\% TA and 95.6\% EA, 18.4, 26.6 and 32.0 points above the strongest baseline, ITP (Figure~\ref{fig:real_radar}). The baselines' scores drop from DA to EA, as they often misplace the new subtasks even when they choose the right decision, whereas CIRRA keeps all three on par. The largest gap is in IC: CIRRA always interrupts at the correct moment (100\%), since the boundary follows directly from the interruptibility mode, whereas the baselines reach at most 51.6\% on average. Queueing (3) is the easiest combination and fusion beyond a navigation step (4) the hardest, where CIRRA's EA drops to 84\% and the baselines reach 40--48\%.

% \paragraph{Results.}
% Averaged over the five combinations, CIRRA reaches 95.6\% DA, 95.8\% TA
% and 95.6\% EA, exceeding the strongest baseline, ITP, by 18.4, 26.6 and
% 32.0 percentage points (pp) (Figure~\ref{fig:real_radar}). The
% baselines' scores drop from DA to TA to EA: they often choose the right
% decision type but misplace the incoming subtasks or alter the remaining
% sequence, whereas CIRRA's rule-constrained candidate generation keeps TA
% and EA on par with DA.
% % TODO (optional): one sentence on why the baseline ranking differs from CHIRP.
% The largest gap is in interruption compliance. CIRRA lets $u_t$ take
% effect at the gold boundary in every episode (100\% IC), since $b_t$ is
% derived deterministically from the interruptibility mode of the active
% subtask, whereas the baselines reach 25--78\% IC across combinations
% (at most 51.6\% on average);
% \ph{dominant baseline failure, e.g., interrupting \sk{BALL\_CAN} or
% waiting unnecessarily at an \texttt{immediate} skill}.
% Queueing (3) is the easiest combination for all methods, while fusion
% beyond a navigation step (4) is the hardest: CIRRA's EA drops to 84\%
% and the baselines reach only 40--48\%.
% \ph{CIRRA's remaining errors in combination 4.}

\section{Limitations}
CHIRP is text-based, and our real-robot study spans two tables and eight skills. Remaining errors stem almost entirely from the 8B compiler rather than the reconciliation mechanism, which is exact under oracle specifications, so CIRRA directly benefits from stronger compilers.

%% file: Tables/main_results_table.tex
% AUTO-GENERATED by benchmark/make_tables.py from results/*_e2e.json.
% Regenerate instead of hand-editing numbers.
\begin{table*}[t]
\centering
\footnotesize
\setlength{\tabcolsep}{3pt}
\begin{tabular}{@{}lcccccccc@{}}
\toprule
Method & Appliance & Clean & Tidy & Cook & Fetch & Waste & Overall & \shortstack{Compile-\\correct} \\
\midrule
LLM-direct & 65 & 80 & 55 & 35 & 45 & 75 & 59.2 & 62.8 \\
\midrule
Conjunctive replan & 40 & 40 & 40 & 40 & 35 & 35 & 38.3 & 55.8 \\
CoT replan~\cite{wei2022chain} & 50 & 40 & 30 & 55 & 35 & 35 & 40.8 & 51.2 \\
LLM-Modulo~\cite{kambhampati2024position} & 40 & 40 & 40 & 40 & 35 & 40 & 39.2 & 58.1 \\
Reflexion~\cite{shinn2023reflexion} & 45 & 35 & 40 & 45 & 50 & 50 & 44.2 & 58.1 \\
ITP~\cite{li2023interactive} & 45 & 40 & 35 & 35 & 45 & 0 & 33.3 & 39.5 \\
CoPAL~\cite{joublin2024copal} & 25 & 20 & 20 & 20 & 40 & 40 & 27.5 & 32.6 \\
CoMuRoS~\cite{borate2025llm} & 40 & 40 & 40 & 30 & 10 & 35 & 32.5 & 44.2 \\
DPPM~\cite{lu2025decompose}$^\dagger$ & 35 & 50 & 50 & 50 & 40 & 5 & 38.3 & 51.2 \\
\midrule
SayCan~\cite{ahn2022can}$^\dagger$ & 40 & 40 & 40 & 40 & 35 & 40 & 39.2 & 58.1 \\
Memory-agent~\cite{glocker2025llm}$^\dagger$ & 50 & 40 & 35 & 45 & 40 & 45 & 42.5 & 58.1 \\
\midrule
Ours w/o judge & \textbf{85} & \textbf{85} & \textbf{70} & \textbf{65} & \textbf{75} & 55 & 72.5 & 81.4 \\
\textbf{Ours (full)} & \textbf{85} & \textbf{85} & \textbf{70} & \textbf{65} & 45 & \textbf{95} & 74.2 & 93.0 \\
\bottomrule
\end{tabular}
\caption{Decision agreement (DA, \%) on the 120 CHIRP episodes. All methods share one Qwen3-8B compilation and differ only in reaction policy; \emph{Compile-correct} covers the 43 correctly compiled episodes. $^\dagger$Operator-level adaptation (Appendix \ref{app:baselines}).}
\label{tab:main-results}
\end{table*}

%% file: Tables/metrics_table.tex
\begin{wraptable}{r}{0.52\textwidth}
\centering
\caption{Fusion performance (\%) on the 40 fuse-gold episodes;
\emph{oracle spec} feeds the ground-truth specification.
$^{\ast}$Conj.\ replan, LLM-Modulo, CoMuRoS,
SayCan$^\dagger$, Memory-agent$^\dagger$.}
\label{tab:metric-suite}

\footnotesize
\setlength{\tabcolsep}{2pt}
\renewcommand{\arraystretch}{0.92}

\begin{tabular*}{\linewidth}{@{\extracolsep{\fill}}lccccc@{}}
\toprule
\multirow{2}{*}{Method}
    & \multicolumn{3}{c}{End-to-end}
    & \multicolumn{2}{c}{Oracle spec} \\
\cmidrule(lr){2-4} \cmidrule(l){5-6}
    & DA & TA & EA & DA & EA \\
\midrule
LLM-direct
    & 27.5 & 22.5 & 22.5 & 100.0 & 100.0 \\
Five baselines$^{\ast}$
    & 0.0 & 0.0 & 0.0 & 0.0 & 0.0 \\
CoT replan
    & 10.0 & 7.5 & 2.5 & 10.0 & 10.0 \\
Reflexion
    & 10.0 & 10.0 & 2.5 & 20.0 & 17.5 \\
ITP
    & 50.0 & 47.5 & 15.0 & 47.5 & 40.0 \\
CoPAL
    & 2.5 & 2.5 & 0.0 & 0.0 & 0.0 \\
DPPM\rlap{\smash{$^\dagger$}}
    & 45.0 & 35.0 & 12.5 & 52.5 & 40.0 \\
Ours w/o judge
    & 72.5 & 72.5 & 47.5 & 100.0 & 100.0 \\
\textbf{Ours (full)}
    & 75.0 & 75.0 & 75.0 & 100.0 & 100.0 \\
\bottomrule
\end{tabular*}
\end{wraptable}
 

%% file: Tables/scale_ablation_table.tex
\begin{wraptable}{r}{0.48\textwidth}
\vspace{-14pt}
\centering
\caption{Compiler-scale ablation (\%). Compiler and judge are swapped together; the symbolic arbitration is fixed. Compile fidelity: episodes compiled exactly as the oracle.}
\label{tab:scale-ablation}
\footnotesize
\setlength{\tabcolsep}{2pt}
\begin{adjustbox}{max width=\linewidth}
\begin{tabular*}{\linewidth}{@{\extracolsep{\fill}}lcccc@{}}
\toprule
Compiler & \shortstack{Compile\\fidelity} & \shortstack{Ours\\DA} & \shortstack{Ours DA\\(comp.-\\corr.)} & \shortstack{LLM-\\direct\\DA} \\
\midrule
Qwen3-4B & 31.7 & 65.8 & 86.8 & 43.3 \\
Qwen3-8B & 35.8 & 74.2 & 93.0 & 59.2 \\
Qwen3-32B & 68.3 & 91.6 & 99.2 & 87.5 \\
\bottomrule
\end{tabular*}
\end{adjustbox}
\vspace{-8pt}
\end{wraptable}

%% file: Tex/conclusion.tex
\section{Conclusion}
We presented CIRRA, a framework for reconciling newly arriving instructions with an ongoing household task. Instead of regenerating the plan, CIRRA inserts new subtasks into the current schedule after a boundary set by the interruptibility of the active subtask, avoiding plan ambiguity; an LLM conflict gate rejects insertions that break task dependencies, preventing logical inconsistency; and new subtasks are fused into segments at the same location, reusing pending navigation to avoid execution redundancy. On CHIRP, CIRRA reaches 74.2\% decision agreement, 30 points above the strongest replanning baseline, and every correct fusion decision yields a correctly placed, conflict-free schedule. On a Unitree G1 humanoid, it interrupts ongoing skills at the correct moment in every trial and outperforms all baselines on every metric. These results show that constraining where an LLM may change a plan makes replanning both more accurate and safer to execute.

%% file: Tex/appendix.tex
% ============================================================================
% Appendix A: the CHIRP benchmark; B: baseline reproduction details;
% C: real-world deployment details. \input this file after the bibliography.
% ============================================================================
\appendix
\section{The CHIRP Benchmark}
\label{app:benchmark}

\subsection{Construction Pipeline}

The benchmark is built in three deterministic stages, all shipped with the
code.

\paragraph{Stage 1: hand-authored catalog.}
Every episode originates from a hand-written catalog entry
that fixes the scenario: the house,
the robot's position and progress inside the ongoing task, the queued
tasks, the injected utterance with two paraphrases, the oracle
specification of the utterance, and the gold decision with its target. The
catalog is organized as six categories of twenty episodes. The backbone of
each category is a set of eight minimal pairs, one per room: the a-member
injects a request whose location coincides with the scene the robot is
currently working in and the b-member injects a nearly identically worded
request anchored elsewhere, so a system that ignores execution state is
guaranteed to miss one member of each pair. In four categories the pairs
carry the plain gold split (a fuse, b queue); the two design decrees below
reshape the remaining two, moving the waste a-members and one fetch
b-member to clarify. Each category adds two preempt episodes (emergencies
or explicit do-this-first orders) and two clarify episodes (underspecified
action or location).

\paragraph{Stage 2: template generator.}
A template generator renders every catalog entry into a
self-contained JSON episode: it expands the task templates into subtask
sequences, stamps the world snapshot, renders the human-readable text
vignette consumed by baselines, and copies the oracle specification and
gold label. Subtasks follow the five-attribute tuple of the paper (skill,
object, location, execution status, interruptibility mode); navigation
subtasks carry no object attribute, and no subtask carries any duration
attribute. All location attributes draw from one normalized room-level
vocabulary (the eight rooms of the house), on both the ongoing tasks and
the oracle specifications, so the method's exact-equality destination
matching is well defined across the whole batch.

\paragraph{Stage 3: validation.}
A schema validator inside the generator checks every episode before it is
written: all twelve axis fields are present and take legal values, gold
decisions are one of the four classes, referenced rooms exist in the
house, oracle subtasks are well-formed, and the no-duration rule holds.
The 120 episodes are byte-reproducible from the catalog plus the
placement stamper described below.

\paragraph{Design rules.}
Two decrees shape the label distribution. First, vague destination words
(``take it out'', ``the usual place'') never resolve to a room: a subtask
location must be a concrete room, so such utterances are ambiguous by
construction and gold clarify at injection. This moves eight
waste-handling episodes and one fetch episode to clarify, which is why
the waste category contains no fuse gold. Second, cooking is pinned to
the kitchen: cooking-context episodes probe whether a system respects a
pinned context rather than surface geometry. Labels were authored by the
benchmark designers; every relabeling applied during construction is recorded in each
episode's \texttt{provenance} field.

\subsection{Episode Schema}

Each episode JSON carries: \texttt{id}; \texttt{house} (the eight-room
layout); \texttt{world} (robot position, user position, object placements,
held items, standing facts); \texttt{active} (the ongoing task: subtask
sequence, progress cursor, in-flight interruptibility mode);
\texttt{queue} (zero or more queued tasks); \texttt{injection} (the
utterance, two human-checked paraphrases, and the oracle specification);
\texttt{gold} (decision, target with its placement anchors, rationale);
\texttt{axes} (twelve
generation axes: decision, phase, urgency, spatial relation, hands, gate,
constraint, contract, category, clarify point, queue depth, stream
length); \texttt{pair} (minimal-pair identifier); \texttt{provenance};
and \texttt{rendering} (the plain-text vignette consumed by baselines).

\subsection{Category Composition}

Table~\ref{tab:category-composition} gives the gold decision composition
of the six categories.

\begin{table}[h]
\centering
\small
\caption{Gold decision composition of the six categories.}
\label{tab:category-composition}
\begin{tabular}{@{}llcccc@{}}
\toprule
Category & Probe axis & Fuse & Queue & Clarify & Preempt \\
\midrule
Appliance & same/diff & 8 & 8 & 2 & 2 \\
Clean & same/diff & 8 & 8 & 2 & 2 \\
Tidy & same/diff & 8 & 8 & 2 & 2 \\
Cook & pinned ctx. & 8 & 8 & 2 & 2 \\
Fetch & shared dest. & 8 & 7 & 3 & 2 \\
Waste & co-carry & 0 & 8 & 10 & 2 \\
\midrule
Total & & 40 & 47 & 21 & 12 \\
\bottomrule
\end{tabular}
\end{table}

\subsection{Placement Targets and Method-Structure Probes}

Beyond the decision label, every fuse gold carries a \texttt{placements}
target: the visit the new behavior must ride and its local position inside
that visit (or a reuse marker when an identical pending subtask already
exists). The decision labels are hand-authored; the placement layer follows
one mechanical doctrine, the frontmost admissible position of the earliest
matched visit, and is stamped by a judge-free symbolic pass and audited by
hand, with the derivation recorded in each episode's \texttt{provenance}.

A twelve-episode probe supplement (released alongside the main batch, not
counted in its tables) covers three mechanism structures the main batch
does not isolate: duplicate reuse (PBR01--PBR04: the requested behavior is
already pending in the matched visit and must be reused, not duplicated),
insertion ahead of an immediate-interruptible in-flight behavior
(PBI01--PBI04), and traversal past an inadmissible earliest visit to a
later same-destination visit, with the undeliverable part of the request
queued as remainder (PBV01--PBV04). The probes are generated and stamped
by the same pipeline as the main batch.

\subsection{Full Episode Listing}

Tables~\ref{tab:episodes-appliance} through \ref{tab:episodes-waste_carry}
list every episode by category: identifier, the original surface form of
the injected utterance, and the gold decision. The two paraphrases, the
world snapshots, the ongoing and queued tasks, the oracle specifications,
and the gold targets are in the released JSON files, one file per episode.

\input{Tables/episode_listing}

\clearpage

% ============================================================================
% Appendix B: baseline reproduction details
% ============================================================================

\section{Baseline Reproduction Details}
\label{app:baselines}

All baselines run inside one harness under identical conditions: the same
shared compilation feeds every plan-consuming method, every LLM call goes
through one gateway with one decoding contract (temperature 0, fixed seed,
thinking disabled, schema-constrained JSON), and every regeneration
baseline emits reference-constrained schedules (\texttt{old:<id>} keeps a
pending step, \texttt{new:<i>} schedules a new one), so the induced
decision and the commitment violations are computed by one deterministic
classifier with no LLM in the grading loop. For each baseline we record
the first-hand source, what is reproduced faithfully, and every
adaptation made to run it on our episodes. Raw source material (original
prompt files where published) ships in \texttt{benchmark/baselines\_ref/}.

\paragraph{ITP~\citep{li2023interactive}.}
Faithful: the replanning context is the task guide plus completed steps
plus the new request plus chat history, continued into the prior
conversation, and the system plans one task thread at a time, as in the
original drink-making deployment. Adaptations: the original emits GPT-4
function calls, ours emits leaf references so the induced decision is
computable; the task guide's few-shot menu examples are replaced by our
domain skill list; the backbone is the protocol-wide Qwen3-8B instead of
GPT-4.

\paragraph{CoPAL~\citep{joublin2024copal}.}
Faithful, rebuilt from the original prompt files in the authors'
repository: the Travi role frame (human, Travi, robot), the four numbered
duties, the ``do not replicate actions already done'' instruction, the
feedback re-prompt block (request, feedback, command history, JSON
state), the output fields Goal / Objects needed / State summary /
Remaining actions, and the Error, Reason, Suggestion triplet phrasing.
Adaptations: the feedback triplet carries a user interjection instead of
an execution error (the original replans on errors, not on new
instructions, which is itself a point of comparison); the four-layer
cognitive architecture collapses to the high-level replanning layer for a
text benchmark; temperature follows the unified protocol instead of the
original 0.8; the replan loop cap becomes one round.

\paragraph{CoMuRoS~\citep{borate2025llm}.}
Faithful: the Task Manager prompt's static block (rules and format) and
dynamic block (user command, chat history, robot state, task states,
events); the three task states COMPLETED / IN\_PROGRESS / INTERRUPTED;
replanning reorders only non-completed tasks; the new instruction arrives
as a detected event; whole-task granularity (tasks are never split),
which is this family's native action space. Adaptations: the multi-robot
allocation axis does not apply (single robot); the paper does not publish
the verbatim template, so the block wording is rebuilt from its
description; output is constrained to a task-order JSON.

\paragraph{DPPM~\citep{lu2025decompose}$^\dagger$.}
Faithful: the published merge instruction (``merge and revise two partial
plans based on feedback from critics into one that meets the
requirements''), the Partial Plan A / Feedback A / Partial Plan B /
Feedback B field structure, the pure-JSON output requirement, and
explicit conflict annotations (transport exclusivity in the original,
hand capacity and forbidden co-carries here). Adaptations are larger,
hence the dagger: the original is offline text travel planning, ours is
an online embodied stream; Plan A comes from the live commitments rather
than a parallel sub-planner; the verify-refine loop (up to ten rounds)
becomes a single merge; the Cartesian candidate combination does not
apply to a two-plan merge.

\paragraph{CoT replan~\citep{wei2022chain}.}
The chain-of-thought baseline of the DPPM comparison pool: identical task
and input to Conjunctive replan, with the schema forcing an in-band
\texttt{reasoning} field before the plan.

\paragraph{LLM-Modulo~\citep{kambhampati2024position}.}
The generate-test baseline of the DPPM pool. The proposal LLM is checked
by a sound verifier, and the verifier is our symbolic simulator
(commitment accounting, location contracts, hand capacity, forbidden
co-carries), with violations fed back for revision, up to three rounds.
This is a deliberate concession: the strongest verifier available in this
setting, isolating ``verifier without enumeration'' from ``enumeration
plus verifier.''

\paragraph{Reflexion~\citep{shinn2023reflexion}.}
Proposal (sharing the Conjunctive replan prompt and cache), one
self-reflection pass prompted to check for dropped steps, infeasible
order, forbidden co-carries, and missed ride-alongs, then one revision.
No external verifier: together with LLM-Modulo this forms the
introspection-versus-external-check contrast.

\paragraph{SayCan~\citep{ahn2022can}$^\dagger$.}
A structural constant, stated openly: SayCan has no clarification, no
emergency handling, and no weaving mechanism; each instruction is planned
independently and executed in arrival order, so in this decision space it
is queue-at-tail on every episode and involves no LLM call. The row
quantifies the ceiling of the sequential one-request contract under the
field's foundational system; its affordance-scoring internals are
invisible to DA and TA and reproducing them would not change any reported
number.

\paragraph{Memory-agent~\citep{glocker2025llm}$^\dagger$.}
Faithful, using the original system message from the authors' repository:
the object-list dependence, the full clarification rules (synonyms pass,
near-matches ask, missing objects report, uncertainty asks), the
single-request task-JSON output contract, and the chat-history mechanism
(the active task enters through it). Its native clarification behavior
makes this row meaningful on clarify episodes. Adaptations, hence the
dagger: a third task type ``action'' is added (the original knows only
placement and handover, without which sixty chore episodes are
inexpressible at the vocabulary level and the comparison would measure
vocabulary, not scheduling); its three native output forms merge into one
schema for constrained decoding; the routing agent is bypassed (all
inputs are task requests); the RAG memory idles because a frozen single
moment has no long interaction history, so its memory axis is not
assessed here; the backbone is the protocol-wide Qwen3-8B.

\paragraph{Considered and excluded.}
LLM+P encodes problems into PDDL for a classical planner, but arbitration
is not goal-reachability planning: commitment order, preemption
semantics, and same-visit structure have no natural PDDL expression, so
forcing the encoding would strawman the method. Code as Policies~\citep{liang2023code} and
ProgPrompt~\citep{singh2022progprompt} belong to the code-generation family (the LLM writes a program
against the task tree), a different comparison family left as an
extension. CoMuRoS's own paper reports no quantitative baseline to
borrow.

\paragraph{Grading fairness.}
Classification and violation counting use the same deterministic
classifier for every regeneration baseline. One known expressiveness
boundary is disclosed: a flat reference sequence cannot express
``ride the same visit'' tree structure, so end-state fuse golds are
systematically classified queue for flat-output systems. This is not a
grading artifact but the direct cost of flattening away commitment
structure, which is part of the paper's argument.

\clearpage
\providecommand{\sk}[1]{\texttt{#1}}
\providecommand{\apprr}{\raggedright\hspace{0pt}}

\section{Real-World Deployment Details}
\label{app:real}
 
\paragraph{Skills.}
Table~\ref{tab:real_skills} lists the eight behavioral skills and the
navigation skill exposed to CIRRA, together with their interruptibility
modes. A skill is \texttt{immediate} if halting it at any time leaves no
object in a hazardous state, \texttt{safe\_point} if it passes through
stable intermediate states, with a safe boundary after each placement or
stage, and \texttt{uninterruptible} if it keeps an object in a precarious
configuration throughout execution. Safe boundaries are detected online
from the gripper state and a per-skill stage counter. A paused skill
switches its policy to standby mode, identically for every method, and
later resumes from the current observation.
 
\begin{table}[!ht]
\centering
\small
\setlength{\tabcolsep}{4pt}
\caption{Real-world skill library. Each visuomotor policy is exposed to
CIRRA as one behavioral skill with an interruptibility mode $\iota$.}
\label{tab:real_skills}
\begin{tabular}{@{}lp{4.0cm}lp{4.2cm}@{}}
\toprule
Skill & Task & $\iota$ & Safe boundary / rationale \\
\midrule
\sk{BALL\_CAN} & \apprr Insert a ball into a hand-held can & \texttt{uninterruptible} & \apprr bimanual; releasing drops the can or ball \tabularnewline
\sk{BASKET} & \apprr Put a block into a basket & \texttt{immediate} & \apprr halting is harmless at any time \tabularnewline
\sk{SWAP} & \apprr Swap the items on two plates & \texttt{safe\_point} & \apprr after each item is placed on a plate \tabularnewline
\sk{DRAWER} & \apprr Put a block into the bottom drawer & \texttt{safe\_point} & \apprr after each stage (open, place, close) \tabularnewline
\sk{SORT\_ITEM} & \apprr Sort items into baskets by color & \texttt{immediate} & \apprr halting is harmless at any time \tabularnewline
\sk{HANG} & \apprr Hang a tape roll on a rack & \texttt{uninterruptible} & \apprr the held tape is aligned with the peg \tabularnewline
\sk{SORT\_TOOL} & \apprr Sort tools into a tool case & \texttt{safe\_point} & \apprr after each tool is placed in its slot \tabularnewline
\sk{STACK\_CUBE} & \apprr Stack cubes & \texttt{safe\_point} & \apprr after each cube is released \tabularnewline
\midrule
\sk{NAV} & \apprr Walk to table $A$ or $B$ & \texttt{immediate} & \apprr walking can be stopped at any time \tabularnewline
\bottomrule
\end{tabular}
\end{table}
 
\paragraph{Task combinations.}
Table $A$ stands to the left of table $B$, and the tasks on each table
are arranged from left to right in the order listed in
Table~\ref{tab:real_scenarios}. No ordering rule is given to the agent,
which must infer the layout from $\mathcal{E}_t$. \sk{SORT\_ITEM} is a
single skill whose item order is chosen by the policy. Each combination
is run 25 times, and in each trial $u_t$ is issued at a different phase
of the ongoing sequence, e.g., during navigation, at the onset of a
grasp, or mid-manipulation; the gold decision, boundary $b_t$ and revised
sequence are annotated for every trial before execution. Trials in which
a manipulation policy fails to grasp are excluded. Ambiguity resolution
is evaluated on CHIRP and not repeated here.
 
\begin{table}[!ht]
\centering
\footnotesize
\setlength{\tabcolsep}{3pt}
\renewcommand{\arraystretch}{1.15}
\caption{Real-world task combinations. Tasks on each table are listed
from left to right. In each of the 25 trials, $u_t$ arrives while a
different subtask of $\mathcal{P}_t$ is active, and the gold boundary
$b_t$ follows from the interruptibility mode of that subtask
(Table~\ref{tab:real_skills}). \sk{SORT\_ITEM}($\cdot$) lists the items
to sort.}
\label{tab:real_scenarios}
\begin{tabular}{@{}p{0.2cm}p{2.15cm}p{2.9cm}p{3.35cm}p{4.3cm}@{}}
\toprule
\# & Layout & Ongoing $\mathcal{P}_t$ & New instruction $u_t$ & Gold decision and $\mathcal{P}'_t$ \\
\midrule
1 &
  \apprr $A$: \sk{SORT\_ITEM} \newline $B$: \sk{BALL\_CAN}, \sk{DRAWER} &
  \apprr \sk{NAV}($A$), \sk{SORT\_ITEM}(red, blue, yellow) &
  \apprr ``Now go to table B, put the tennis ball into the can, then put the block into the bottom drawer.'' &
  \apprr \emph{preempt} at $b_t$: \sk{NAV}($B$), \sk{BALL\_CAN}, \sk{DRAWER}, then \sk{NAV}($A$) and the unfinished part of $\mathcal{P}_t$ \tabularnewline
\addlinespace
2 &
  \apprr $A$: \sk{BALL\_CAN} \newline $B$: \sk{BASKET}, \sk{SORT\_TOOL} &
  \apprr \sk{NAV}($A$), \sk{BALL\_CAN} &
  \apprr ``Now go and put the screwdriver on table B into the tool case.'' &
  \apprr \emph{preempt} at $b_t$: \sk{NAV}($B$), \sk{SORT\_TOOL}, then \sk{NAV}($A$) and the unfinished part of $\mathcal{P}_t$, if any \tabularnewline
\addlinespace
3 &
  \apprr $A$: \sk{HANG}, \sk{SWAP} \newline $B$: \sk{SORT\_TOOL} &
  \apprr \sk{NAV}($A$), \sk{HANG}, \sk{SWAP} &
  \apprr ``By the way, put the tools on table B into the tool case.'' &
  \apprr \emph{queue}: \sk{NAV}($A$), \sk{HANG}, \sk{SWAP}, \sk{NAV}($B$), \sk{SORT\_TOOL} \tabularnewline
\addlinespace
4 &
  \apprr $A$: \sk{SORT\_ITEM} \newline $B$: \sk{BALL\_CAN}, \sk{DRAWER} &
  \apprr \sk{NAV}($B$), \sk{BALL\_CAN}, \sk{NAV}($A$), \sk{SORT\_ITEM}(red), \sk{NAV}($B$), \sk{DRAWER} &
  \apprr ``By the way, put the yellow and blue items into the baskets of the same colors.'' &
  \apprr \emph{fuse} into the $A$ segment: \sk{NAV}($B$), \sk{BALL\_CAN}, \sk{NAV}($A$), \sk{SORT\_ITEM}(red), \sk{SORT\_ITEM}(yellow, blue), \sk{NAV}($B$), \sk{DRAWER} \tabularnewline
\addlinespace
5 &
  \apprr $A$: \sk{BASKET} \newline $B$: \sk{HANG}, \sk{DRAWER} &
  \apprr \sk{NAV}($A$), \sk{BASKET}, \sk{NAV}($B$), \sk{DRAWER} &
  \apprr ``By the way, hang the tape on table B on the rack.'' &
  \apprr \emph{fuse} into the $B$ segment before \sk{DRAWER} (left-to-right layout): \sk{NAV}($A$), \sk{BASKET}, \sk{NAV}($B$), \sk{HANG}, \sk{DRAWER} \tabularnewline
\bottomrule
\end{tabular}
\end{table}

%% file: Tables/episode_listing.tex
% AUTO-GENERATED by benchmark/make_episode_listing.py.
% Full listing of the 120 home8 episodes, split into six
% per-category table* blocks (two-column class compatible).
\begin{table*}[t]
\centering
\footnotesize
\setlength{\tabcolsep}{4pt}
\begin{tabular}{@{}lp{13.2cm}l@{}}
\toprule
ID & Utterance & Gold \\
\midrule
HA01a & Switch off the bedside lamp too while you're in there. & fuse \\
HA01b & Turn on the exhaust fan in the bathroom too. & queue \\
HA02a & Power the printer down too while you're at the desk. & fuse \\
HA02b & Close the garage door too. & queue \\
HA03a & Switch the range hood on as well while you're there. & fuse \\
HA03b & Latch the living room window too. & queue \\
HA04a & Turn the towel warmer on too while you're in there. & fuse \\
HA04b & Switch off the desk lamp in the study too. & queue \\
HA05a & Turn the floor lamp on too while you're by the window. & fuse \\
HA05b & Turn the garden sprinkler on out in the yard too. & queue \\
HA06a & Check the breaker box too while you're down there. & fuse \\
HA06b & Start the dishwasher in the kitchen too. & queue \\
HA07a & Kill the garage lights too while you're there. & fuse \\
HA07b & Check the water heater in the basement too. & queue \\
HA08a & Turn the yard lights on too while you're out there. & fuse \\
HA08b & Close the bedroom window too. & queue \\
HA09 & The smoke alarm is going off in the kitchen - go check right now! & preempt \\
HA10 & The garage door is closing on the car - stop it now! & preempt \\
HA11 & Turn off the lamp too. & clarify \\
HA12 & Turn on the heater, it's getting cold. & clarify \\
\bottomrule
\end{tabular}
\caption{Episode listing (1 of 6): the Appliance category (20 episodes). The utterance is the original surface form.}
\label{tab:episodes-appliance}
\end{table*}

\begin{table*}[t]
\centering
\footnotesize
\setlength{\tabcolsep}{4pt}
\begin{tabular}{@{}lp{13.2cm}l@{}}
\toprule
ID & Utterance & Gold \\
\midrule
HC01a & Wipe the windowsill in here too while you're at it. & fuse \\
HC01b & The garage floor needs sweeping too. & queue \\
HC02a & Give the desk a wipe too while you're dusting. & fuse \\
HC02b & The stove top in the kitchen needs a wipe too. & queue \\
HC03a & Do the stove top as well while you're wiping. & fuse \\
HC03b & The bathroom mirror needs wiping as well. & queue \\
HC04a & Wipe the mirror too while you're in there. & fuse \\
HC04b & The garden table outside needs a wipe too. & queue \\
HC05a & Do the TV stand too while you're wiping. & fuse \\
HC05b & The basement floor needs sweeping too. & queue \\
HC06a & Wipe the shelves down there too while you're at it. & fuse \\
HC06b & The desk in the study needs dusting too. & queue \\
HC07a & Wipe the workbench too while you're in there. & fuse \\
HC07b & The bedroom windowsill needs wiping too. & queue \\
HC08a & Wipe the garden table too while you're out there. & fuse \\
HC08b & The living room table needs a wipe too. & queue \\
HC09 & The kids spilled juice all over the living room - clean it up now! & preempt \\
HC10 & The bathtub is overflowing - turn off the tap now! & preempt \\
HC11 & Wipe down the table too. & clarify \\
HC12 & Clean up the mess when you can. & clarify \\
\bottomrule
\end{tabular}
\caption{Episode listing (2 of 6): the Clean category (20 episodes). The utterance is the original surface form.}
\label{tab:episodes-clean}
\end{table*}

\begin{table*}[t]
\centering
\footnotesize
\setlength{\tabcolsep}{4pt}
\begin{tabular}{@{}lp{13.2cm}l@{}}
\toprule
ID & Utterance & Gold \\
\midrule
HT01a & The scarf on the bed goes into the wardrobe too. & fuse \\
HT01b & The book on the study desk needs to go back on the shelf too. & queue \\
HT02a & File the documents into the drawer too while you're there. & fuse \\
HT02b & The cushion in the living room goes back on the sofa too. & queue \\
HT03a & The mug on the counter goes in the cupboard too. & fuse \\
HT03b & The ball out in the yard goes back in the toy bin too. & queue \\
HT04a & Put the shampoo back on its shelf too while you're in there. & fuse \\
HT04b & The flashlight in the basement goes back in its drawer too. & queue \\
HT05a & The magazine goes back on the rack too while you're at the sofa. & fuse \\
HT05b & The extension cord in the garage needs coiling back onto its shelf too. & queue \\
HT06a & Put the flashlight back in its drawer too while you're down there. & fuse \\
HT06b & The scarf on the bed upstairs goes into the wardrobe too. & queue \\
HT07a & Coil the extension cord back onto its shelf too while you're there. & fuse \\
HT07b & The clean dishes on the counter go into the cupboard too. & queue \\
HT08a & The garden gloves go back in the shed box too while you're out there. & fuse \\
HT08b & The towel in the bathroom needs hanging back on the rack too. & queue \\
HT09 & The stove is still on - turn it off first! & preempt \\
HT10 & Grandpa slipped in the bathroom - go check on him now! & preempt \\
HT11 & Put my cup away too. & clarify \\
HT12 & Stash the spare keys somewhere safe. & clarify \\
\bottomrule
\end{tabular}
\caption{Episode listing (3 of 6): the Tidy category (20 episodes). The utterance is the original surface form.}
\label{tab:episodes-tidy}
\end{table*}

\begin{table*}[t]
\centering
\footnotesize
\setlength{\tabcolsep}{4pt}
\begin{tabular}{@{}lp{13.2cm}l@{}}
\toprule
ID & Utterance & Gold \\
\midrule
HK01a & Put the kettle on too while you're at the stove. & fuse \\
HK01b & Also bring my glasses from the bedroom. & queue \\
HK02a & Toast a slice of bread as well while you're there. & fuse \\
HK02b & Also bring my laptop from the study. & queue \\
HK03a & Fill the water pitcher too while you're at the sink. & fuse \\
HK03b & The towel in the bathroom needs hanging back up too. & queue \\
HK04a & Warm up some milk as well while you're brewing. & fuse \\
HK04b & The living room table needs a wipe too. & queue \\
HK05a & Pour a glass of juice too while you're at it. & fuse \\
HK05b & Also bring the storage box up from the basement. & queue \\
HK06a & Rinse the salad greens too while the pot is going. & fuse \\
HK06b & The bike pump in the garage goes back on its hook too. & queue \\
HK07a & Rinse the berries for dessert too while you're mixing. & fuse \\
HK07b & The patio outside needs sweeping too. & queue \\
HK08a & Pour two glasses of water too while you're there. & fuse \\
HK08b & The cushion in the living room goes back on the sofa too. & queue \\
HK09 & The baby is crying upstairs - go check on her right now! & preempt \\
HK10 & Something's burning in the oven - pull it out now! & preempt \\
HK11 & Add my usual sauce. & clarify \\
HK12 & Grab the special seasoning my mom left. & clarify \\
\bottomrule
\end{tabular}
\caption{Episode listing (4 of 6): the Cook category (20 episodes). The utterance is the original surface form.}
\label{tab:episodes-cook}
\end{table*}

\begin{table*}[t]
\centering
\footnotesize
\setlength{\tabcolsep}{4pt}
\begin{tabular}{@{}lp{13.2cm}l@{}}
\toprule
ID & Utterance & Gold \\
\midrule
HF01a & While you're in the bedroom, grab my phone charger too. & fuse \\
HF01b & Also bring the bike pump from the garage. & queue \\
HF02a & Bring the documents from the study along as well. & fuse \\
HF02b & And bring me an apple from the kitchen. & queue \\
HF03a & Grab a snack bar too while you're in the kitchen. & fuse \\
HF03b & Also fetch my phone charger from the bedroom. & queue \\
HF04a & Bring the hair dryer too while you're in the bathroom. & fuse \\
HF04b & Also bring the watering can in from the yard. & queue \\
HF05a & Take the magazine on the couch up too. & fuse \\
HF05b & Also bring the toolbox up from the basement. & clarify \\
HF06a & Bring the spare bulbs up too while you're down there. & fuse \\
HF06b & Also bring my laptop from the study. & queue \\
HF07a & Grab the extension cord too while you're in the garage. & fuse \\
HF07b & Also bring a towel from the bathroom. & queue \\
HF08a & Bring the garden gloves in too while you're out there. & fuse \\
HF08b & Also bring the flashlight up from the basement. & queue \\
HF09 & Quick, I need my medicine right now! & preempt \\
HF10 & Water's leaking under the bathroom sink - go shut the valve now! & preempt \\
HF11 & Bring me my mug. & clarify \\
HF12 & Bring me my headphones. & clarify \\
\bottomrule
\end{tabular}
\caption{Episode listing (5 of 6): the Fetch category (20 episodes). The utterance is the original surface form.}
\label{tab:episodes-fetch}
\end{table*}

\begin{table*}[t]
\centering
\footnotesize
\setlength{\tabcolsep}{4pt}
\begin{tabular}{@{}lp{13.2cm}l@{}}
\toprule
ID & Utterance & Gold \\
\midrule
HW01a & The recycling bag in there goes out too. & clarify \\
HW01b & Bring my pillow down to the living room while you're at it. & queue \\
HW02a & The study recycling goes out too. & clarify \\
HW02b & Also bring the signed documents to the living room for me. & queue \\
HW03a & The kitchen recycling goes out too. & clarify \\
HW03b & Grab me an apple on your way too. & queue \\
HW04a & The bathroom recycling goes out too. & clarify \\
HW04b & Bring the fresh towel to the living room for me too. & queue \\
HW05a & The recycling stack here goes out too. & clarify \\
HW05b & Take the cushion over to the bedroom for me while you're up. & queue \\
HW06a & The basement recycling goes out too. & clarify \\
HW06b & Bring the water bottles up to the kitchen too. & queue \\
HW07a & The garage recycling goes out to the curb too. & clarify \\
HW07b & Bring the soda crate into the kitchen too. & queue \\
HW08a & The yard recycling goes in too. & clarify \\
HW08b & Bring the picked tomatoes into the kitchen too. & queue \\
HW09 & The delivery person is waiting at the gate - go now! & preempt \\
HW10 & The trash truck is on our street - get the bins to the curb right now! & preempt \\
HW11 & Toss the old box too. & clarify \\
HW12 & Throw away the broken umbrella. & clarify \\
\bottomrule
\end{tabular}
\caption{Episode listing (6 of 6): the Waste category (20 episodes). The utterance is the original surface form.}
\label{tab:episodes-waste_carry}
\end{table*}